\documentclass{article}

\RequirePackage[OT1]{fontenc}
\RequirePackage{amsthm,amsmath,amssymb,caption,graphicx,subcaption,tikz}
\RequirePackage{natbib}
\RequirePackage[colorlinks,citecolor=blue,urlcolor=blue]{hyperref}

\definecolor{forestgreen}{rgb}{0.13, 0.55, 0.13}

\def\l{\left}\def\r{\right}
\def\iid{\stackrel{\mbox{\scriptsize i.i.d.}}{\sim}}
\def\K{\mathcal{K}}
\def\mI{\mathbb{I}}\def\mP{\mathbb{P}}\def\mR{\mathbb{R}}
\def\vc{\mathbf{c}}\def\vx{\mathbf{x}}
\def\MA{\mathbf{A}}\def\MB{\mathbf{B}}\def\MC{\mathbf{C}}\def\MH{\mathbf{H}}
\def\MI{\mathbf{I}}\def\MM{\mathbf{M}}\def\MP{\mathbf{P}}
\def\MU{\mathbf{U}}\def\MV{\mathbf{V}}\def\MX{\mathbf{X}}\def\MY{\mathbf{Y}}
\def\MSigma{\mathbf{\Sigma}}
\def\HR{{\mbox{\scriptsize HR}}}

\title{Nuclear penalized multinomial regression with an application to
predicting at bat outcomes in baseball}
\author{Scott Powers, Trevor Hastie and Robert Tibshirani}

\begin{document}

\maketitle

\begin{abstract}
We propose the nuclear norm penalty as an alternative to the ridge penalty for
regularized multinomial regression. This convex relaxation of reduced-rank
multinomial regression has the advantage of leveraging underlying structure
among the response categories to make better predictions. We apply our method,
nuclear penalized multinomial regression (NPMR), to Major League Baseball
play-by-play data to predict outcome probabilities based on batter-pitcher
matchups. The interpretation of the results meshes well with subject-area
expertise and also suggests a novel understanding of what differentiates
players.
\end{abstract}

\section{Introduction}
\label{sec-intro}

A baseball game comprises a sequence of matchups between one batter and one
pitcher. Each matchup, or {\em plate appearance} (PA), results in one of
several outcomes. Disregarding some obscure possibilities, we consider nine
categories for PA outcomes: flyout (F), groundout (G), strikeout (K),
base on balls (BB), hit by pitch (HBP), single (1B), double (2B), triple (3B)
and home run (HR).

A problem which has received a prodigious amount of attention in
sabermetric (the study of baseball statistics)
literature is determining the value of each of the above outcomes, as it leads
to scoring runs and winning games. But that is only half the battle.
Much less work in this field focuses on an equally important problem: optimally
estimating the probabilities with which each batter and pitcher will produce
each PA outcome. Even for ``advanced metrics'' this second task is
usually done by taking simple empirical proportions, perhaps shrinking them
toward a population mean using a Bayesian prior.

In statistics literature, on the other hand, many have developed shrinkage
estimators for a set of probabilities with application to batting averages,
starting with Stein's estimator \citep{EfronMorris75}. Since then, Bayesian
approaches to this problem have been popular. \citet{Morris83} and
\citet{Brown08} used empirical Bayes for estimating batting averages, which are
binomial probabilities. We are interested in estimating multinomial
probabilities, like the nested Dirichlet model of \citet{Null09} and the
hierarchical Bayesian model of \citet{Albert16}. What all of the above works
have in common is that they do not account for the ``strength of schedule''
faced by each player: How skilled were his opponents?

The state-of-the-art approach, Deserved Run Average \citep[DRA]{DRA},
is similar to the adjusted plus-minus model from basketball and the
Rasch model used in psychometrics. The latter models the probability
(on the logistic scale) that a student correctly answers an exam question as
the difference between the student's skill and the difficulty of the question.
DRA models players' skills as random effects and also includes fixed effects
like the identity of the ballpark where the PA took place. Each category of the
response has its own binomial regression model. Taking HR as an example, each
batter $B$ has a propensity $\beta_B^\HR$ for hitting home runs, and each
pitcher $P$ has a propensity $\gamma_P^\HR$ for allowing
home runs. Distilling the model to its elemental form, if $Y$ denotes the
outcome of a PA between batter $B$ and pitcher $P$,
$$\mP(Y = \mbox{HR}|B, P) = \frac{e^{\alpha^\HR + \beta^\HR_B + \gamma^\HR_P}}
    {1 + e^{\alpha^\HR + \beta^\HR_B + \gamma^\HR_P}}.$$
(Actually, in detail DRA uses the probit rather than the logit
link function.)

One bothersome aspect of DRA is that the probability estimates do not sum to
one; a natural solution is to use a single multinomial regression model
instead of several independent binomial regression models.
Making this adjustment would result in a model very similar to ridge
multinomial regression (described in Section~\ref{sub-review}), and we will
compare the results of our model with the results of ridge regression as a
proxy for DRA. Ridge multinomial regression applied to this problem has
about 8,000 parameters to estimate (one for each outcome for each player)
on the basis of about 160,000 PAs in a season,
bound together only by the restriction that probability estimates sum to one.
One may seek to exploit the
structure of the problem to obtain better estimates, as in ordinal regression.
The categories have an ordering, from least to most valuable to the batting
team:

\begin{center}
K $<$ G $<$ F $<$ BB $<$ HBP $<$ 1B $<$ 2B $<$ 3B $<$ HR,
\end{center}
with the ordering of the first three categories depending on the game
situation. But the proportional odds model used for ordinal regression assumes
that when one outcome is more likely to occur, the outcomes close to it in the
ordering are also more likely to occur. That assumption is woefully off-base in
this setting because as we show below, players who hit a lot of home runs tend
to strike out often, and they tend not to hit many triples. The proportional
odds model is better suited for response variables on the Likert scale
\citep{Likert32}, for example.

\begin{figure}[h]
\caption{\it Illustration of the hierarchical structure among the PA outcome
    categories, adapted from \citet{TheSabermetricRevolution}. Blue terminal
    nodes correspond to the nine outcome categories in the data. Orange
    internal nodes have the following meaning: TTO, three true outcomes;
    BIP, balls in play; W, walks; H, hits; O, outs. Outcomes close to each
    other (in terms of number of edges separating them) are likely to co-occur.}
\label{fig-graph}
\centering
\begin{tikzpicture}[->,thick,
    leaf/.style={circle,draw,minimum size=1cm,fill=blue!20},
    split/.style={circle,draw,minimum size=1cm,fill=orange!20},
    head/.style={circle,draw,minimum size=1cm}]
\node[leaf] (BB) at (0, 0) {BB};
\node[leaf] (HBP) at (1.5, 0) {HBP};
\node[leaf] (3B) at (3.25, 0) {3B};
\node[leaf] (2B) at (4.5, 0) {2B};
\node[leaf] (1B) at (5.75, 0) {1B};
\node[leaf] (G) at (7.25, 0) {G};
\node[leaf] (F) at (8.75, 0) {F};
\node[leaf] (K) at (-0.75, 1.5) {K};
\node[split] (W) at (0.75, 1.5) {W};
\node[leaf] (HR) at (2.25, 1.5) {HR};
\node[split] (H) at (4.5, 1.5) {H};
\node[split] (O) at (8, 1.5) {O};
\node[split] (TTO) at (0.75, 3) {TTO};
\node[split] (BIP) at (6.25, 3) {BIP};
\node[head] (PA) at (3.5, 4.5) {PA};
\path
    (PA) edge (TTO)
    (PA) edge (BIP)
    (TTO) edge (K)
    (TTO) edge (W)
    (TTO) edge (HR)
    (BIP) edge (H)
    (BIP) edge (O)
    (W) edge (BB)
    (W) edge (HBP)
    (H) edge (3B)
    (H) edge (2B)
    (H) edge (1B)
    (O) edge (G)
    (O) edge (F);
\end{tikzpicture}
\end{figure}
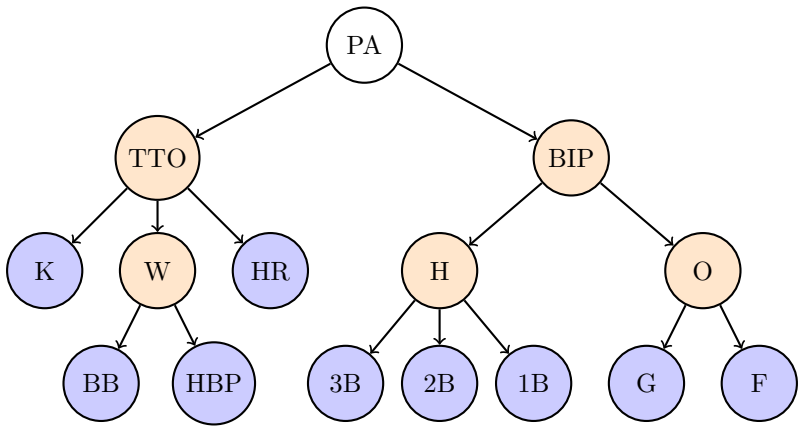

The actual relationships among the outcome categories are more similar to the
hierarchical structure illustrated by Figure~\ref{fig-graph}.
The outcomes fall into two categories:
balls in play (BIP) and the ``three true outcomes'' (TTO). The three true
outcomes, as they have become traditionally known in sabermetric literature,
include home runs, strikeouts and walks (which itself includes BB and HBP). The
distinction between BIP and TTO is important because the former category
involves all eight position players in the field on defense whereas the latter
category involves only the batter and the pitcher. Figure~\ref{fig-graph} has
been designed (roughly) by baseball experts so that terminal nodes close to
each other (by the number of edges separating them) are likely to co-occur.
Players who hit a lot of home runs tend to strike out a lot, and the
outcomes K and HR are only two edges away from each other. Hence the graph
reveals something of the underlying structure in the outcome categories.

\begin{figure}[h]
	\centering
	\begin{subfigure}{.45\linewidth}
		\centering
		\includegraphics[width=\linewidth]{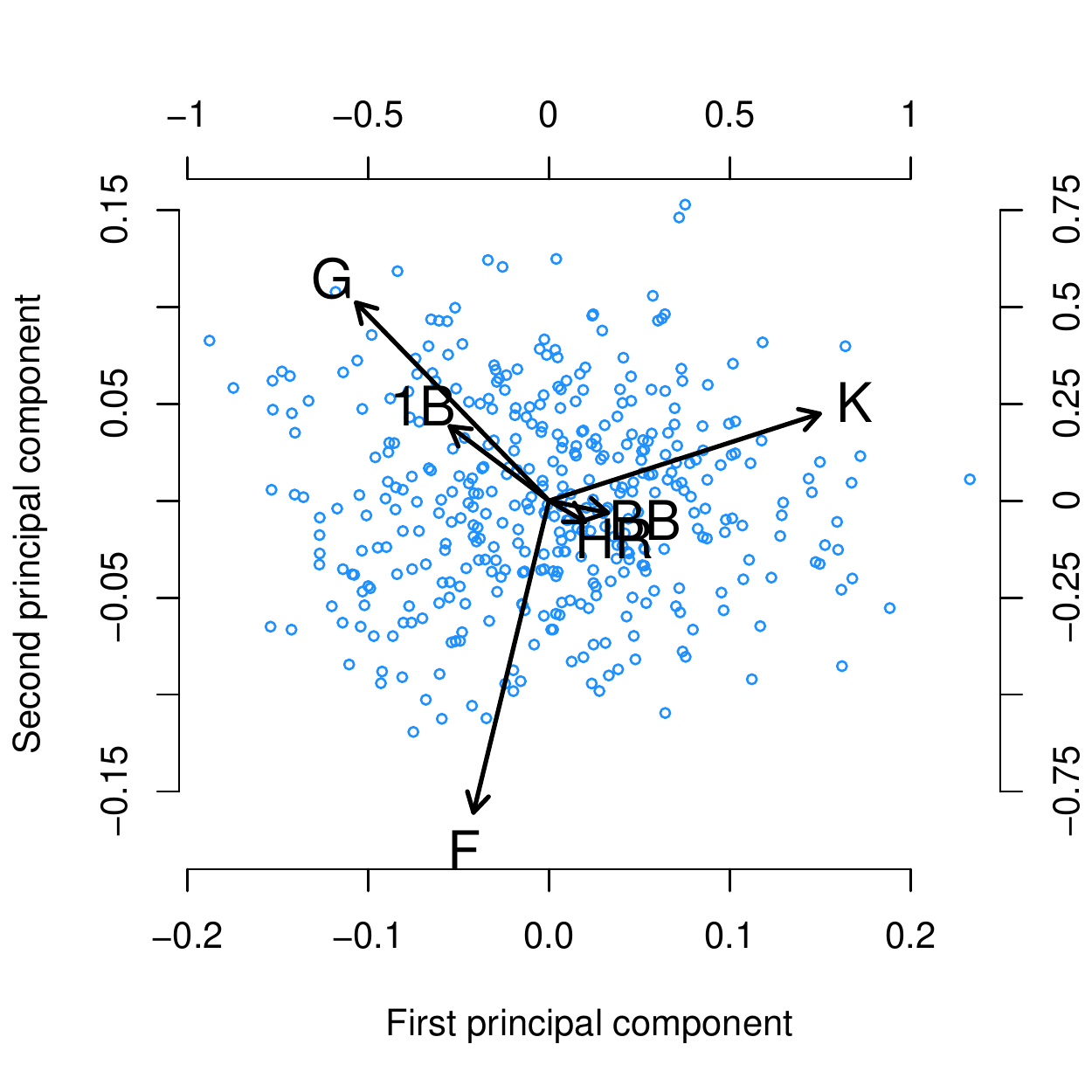}
		\caption{\it Batters}
	\end{subfigure}
	\hspace{1cm}
	\begin{subfigure}{.45\linewidth}
		\centering
		\includegraphics[width=\linewidth]{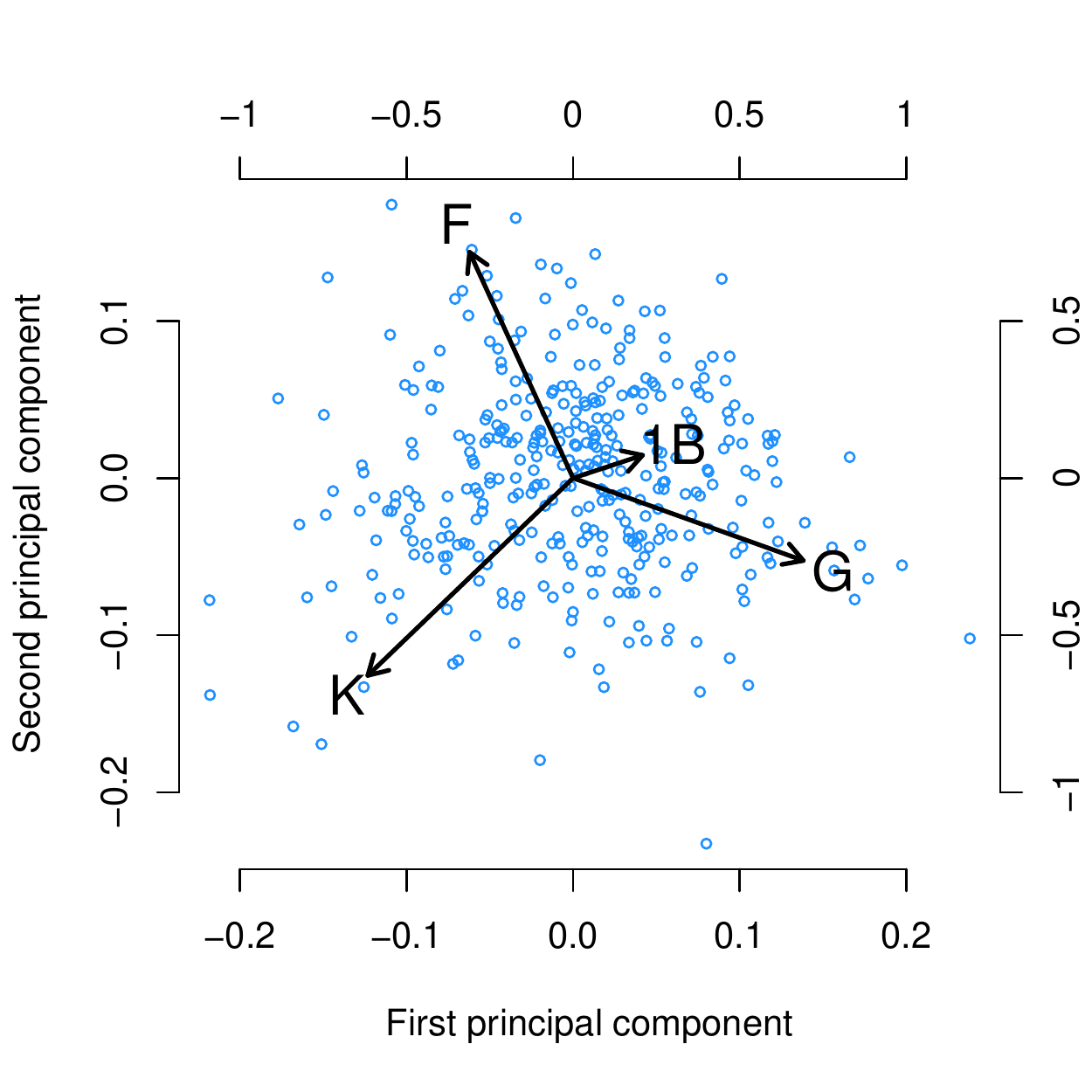}
		\caption{\it Pitchers}
	\end{subfigure}
	\caption{\it Biplots of the principal component analyses of player outcome
    matrices, separate for batters and pitchers. The blue dots represent
    players, and the black arrows (corresponding to the top and right axes)
    show the loadings for the first two principal components on each of the
    outcomes. We exclude outcomes for which the loadings of both of the first
    two principal components are sufficiently small.}
	\label{fig-pca-biplot}
\end{figure}

\begin{figure}[h]
	\centering
	\begin{subfigure}{\linewidth}
		\centering
		\includegraphics[width=.8\linewidth]{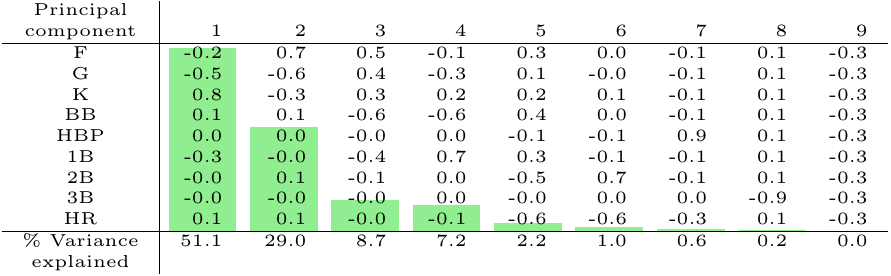}
		\caption{\it Principal components of batter outcome matrix}
	\end{subfigure}
	\hspace{1cm}
	\begin{subfigure}{\linewidth}
		\centering
		\includegraphics[width=.8\linewidth]{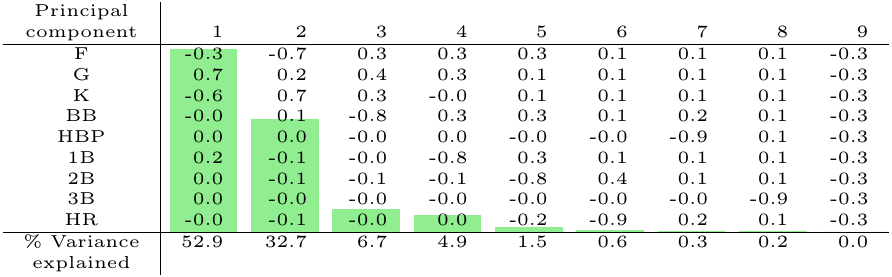}
		\caption{\it Principal components of pitcher outcome matrix}
	\end{subfigure}
	\caption{\it Visualization of principal component analysis of player outcome
        matrices, separate for batters and for pitchers. The player outcome
        matrix has a row for each player giving the proportion of PAs which
        have resulted in each of the nine outcomes in the dataset. The
        visualization shows the loadings for each PC, along
        with a green bar plot corresponding to the percentage of variance
        explained by each PC, which is also printed in the row below the matrix
        of PC loadings.}
	\label{fig-pca}
\end{figure}

This structure is further evidenced by principal component analysis of the
player-outcome matrix, illustrated in Figures~\ref{fig-pca-biplot} and
\ref{fig-pca}. For batters, the
principal component (PC) which describes most of the variance in observed
outcome probabilities has negative loadings on all of the BIP outcomes and
positive loadings on all of the TTO outcomes. For both batters and pitchers,
the percentage of variance explained after two PCs drops off precipitously.

Principal component analysis is useful for illustrating the relationships
between the outcome categories. For example Figure~\ref{fig-pca}(a) suggests
that batters who tend to hit singles (1B) more than average also tend to ground
out (G) more than average. So an estimator of a batter's groundout rate
could benefit from taking into account the batter's single rate, and {\it vice
versa}. This is an example of the type of structure in outcome categories that
motivates our proposal, which aims to leverage this structure to obtain better
regression coefficient estimates in multinomial regression.

In Section \ref{sec-rrmlm} we review reduced-rank multinomial regression, a
first attempt at leveraging this structure. We improve on this in Section
\ref{sec-npmr} by proposing nuclear penalized multinomial regression, a convex
relaxation of the reduced rank problem. We compare our method with ridge
regression in a simulation study in Section \ref{sec-sim}.
In Section \ref{sec-results} we apply our
method and intepret the results on the baseball data, as well as another 
application. The
manuscript concludes with a discussion in Section \ref{sec-discussion}.

\section{Multinomial logistic regression and reduced rank}
\label{sec-rrmlm}

Suppose that we observe data $\vx_i \in \mR^p$ and $Y_i \in \{1, ..., K\}$ for
$i = 1, ..., n$. We use $\MX$ to denote the matrix with rows $\vx_i$,
specifically $\MX = (\vx_1, ..., \vx_n)^T$. The multinomial logistic
regression model assumes that the $Y_i$ are, conditional on $\MX$, independent,
and that for
$k = 1, ..., K:$
\begin{equation}
\label{eqn-multinomial}
\mP(Y_i = k|\vx_i) = \frac{e^{\alpha_k + \vx_i^T\beta_k}}
    {\sum_{\ell = 1}^Ke^{\alpha_\ell + \vx_i^T\beta_\ell}},
\end{equation}
were $\alpha_k \in \mR$ and $\beta_k \in \mR^p$ are fixed, unknown parameters.
The model (\ref{eqn-multinomial}) is not identifiable because an equal increase
in the same element of each of the $\beta_k$ (or in each of the $\alpha_k$)
does not the change the estimated probabilities. That is, for each choice of
parameter values there is an infinite set of choices which have the same
likelihood as the original choice, for any observed data. This problem may
readily be resolved by adopting the restriction for some
$k_0 \in \{1, ..., K\}$ that $\alpha_{k_0} = 0$ and $\beta_{k_0} = \vec 0_p$.
However, depending on the method used to fit the model, this identifiability
issue
may not interfere with the existence of a unique solution; in such a case, we
do not adopt this restriction. See the appendix for a detailed
discussion.

In contrast with logistic regression, multinomial regression involves
estimating not a vector but a matrix of regression coefficients: one for each
independent variable, for each class. We denote this matrix by
$\MB = (\beta_1, ..., \beta_K)$. Motivated by the principal component
analysis from Section \ref{sec-intro}, instead of learning a
coefficient vector for each class, we might do better by learning a
coefficient vector for each of a smaller number $r$ of latent variables, each
having a loading on the classes. For $r = 1$, this is the
{\it stereotype model} originally proposed by \cite{Anderson84}, who observed
its applicability to multinomial
regression problems with structure between the response categories, including
ordinal structure. \cite{Greenland94} argued for the stereotype model as an
alternative in medical
applications to the standard techniques for ordinal categorical regression:
the cumulative-odds and continuation-ratio models.

\cite{YeeHastie03}
generalized the model to reduced-rank vector generalized linear models.
In detail, the reduced-rank multinomial logistic model (RR-MLM) fits
(\ref{eqn-multinomial}) by solving, for some $r \in \{1, ..., K - 1\}$, the
optimization problem:
\begin{align}\begin{split}
\label{eqn-rrmlm}
\underset{\alpha\in\mR^K,~\MB\in\mR^{p\times K}}{\mbox{minimize}}&
-\sum_{i=1}^n\log\l(\sum_{k = 1}^K\frac{e^{\alpha_k + \vx_i^T\beta_k}}
  {\sum_{\ell = 1}^Ke^{\alpha_\ell + \vx_i^T\beta_\ell}}\mI_{\{Y_i = k\}}\r)\\
\mbox{subject to}\hspace{4mm} & \mbox{rank}(\MB) \le r\\
                              & \alpha_1 = 0, \beta_1 = \vec0_p.
\end{split}\end{align}
If rank$(\MB) < r$, then there exist $\MA \in \mR^{p\times r}$,
$\MC \in \mR^{K\times r}$ such that $\MB = \MA\MC^T$. Under this factorization,
the $r$ columns of $\MC$ can be interpreted as defining latent outcome
variables, each with a
loading on each of the $K$ outcome classes. The $r$ columns of $\MA$
give regression coefficient vectors for these latent outcome variables, rather
than the outcome classes.

The optimization problem (\ref{eqn-rrmlm}) is not  convex because
rank$(\cdot)$ is not a convex function. \cite{VGAM} implemented an
alternating algorithm to solve it in the R \citep{R} package {\tt VGAM}.
However this
algorithm is too slow for feasible application to datasets as large as the one
motivating Section \ref{sec-intro} ($n = 176559$, $p = 796$, $K = 9$).

\section{Nuclear penalized multinomial regression}
\label{sec-npmr}

Because of the computational difficulty of solving (\ref{eqn-rrmlm}), we
propose replacing the rank restriction with a restriction on the nuclear norm
$||\cdot||_*$ (defined below) of the regression coefficient matrix.
For some $\rho > 0$, this convex optimization problem is:
\begin{align}\begin{split}
\label{eqn-op}
\underset{\alpha\in\mR^K,~\MB\in\mR^{p\times K}}{\mbox{minimize}}&
-\sum_{i=1}^n\log\l(\sum_{k = 1}^K\frac{e^{\alpha_k + \vx_i^T\beta_k}}
  {\sum_{\ell = 1}^Ke^{\alpha_\ell + \vx_i^T\beta_\ell}}\mI_{\{Y_i = k\}}\r)\\
\mbox{subject to}\hspace{4mm} & ||\MB||_* \le \rho
\end{split}\end{align}

We prefer to frame the problem in its equivalent Lagrangian form: For
some $\lambda > 0$,

\begin{align}
\begin{split}
\label{eqn-npmr}
(\alpha^*, \MB^*) &= \underset{\alpha\in\mR^K,~\MB\in\mR^{p\times K}}{\arg\min}
-\sum_{i=1}^n\log\l(\sum_{k = 1}^K\frac{e^{\alpha_k + \vx_i^T\beta_k}}
    {\sum_{\ell = 1}^Ke^{\alpha_\ell + \vx_i^T\beta_\ell}}\mI_{\{Y_i = k\}}\r)
    + \lambda||\MB||_*\\
    & \equiv \underset{\alpha\in\mR^K,~\MB\in\mR^{p\times K}}{\arg\min}
    -\ell(\alpha, \MB; \MX, Y) + \lambda||\MB||_*
\end{split}
\end{align}

This optimization problem (\ref{eqn-npmr}) is what we call nuclear penalized
multinomial regression (NPMR). We use $\ell(\alpha, \MB; \MX, Y)$ to denote the
log-likelihood of the
regression coefficients $\alpha$ and $\MB$ given the data $\MX$ and $Y$.
The nuclear norm of a matrix is defined as the sum of its singular
values, that is, the $\ell_1$-norm of its vector of singular
values. Explicitly, consider the singular value decomposition of $\MB$
given by $\MU\MSigma\MV^T$, with $\MU \in \mR^{p\times p}$ and
$\MV \in \mR^{K\times K}$ orthogonal and $\MSigma\in\mR^{p\times K}$ having
values $\sigma_1, ..., \sigma_{\min\{p, K\}}$ along the main diagonal and zeros
elsewhere. Then

\begin{equation*}
||\MB||_* = \sum_{d=1}^{\min\{p, K\}}\sigma_d.
\end{equation*}

In the same way that the lasso induces sparsity of the estimated
coefficients in a regression, solving 
(\ref{eqn-npmr}) drives some of the singular values to exactly zero. Because
the number of nonzero singular values is the rank of a matrix, the result is
that the estimated coefficient matrix $\MB^*$ tends to have less than
full rank. Thus (\ref{eqn-npmr}) is a convex relaxation of the reduced-rank
multiomial logistic regression problem, in much the same way as the lasso is a
convex relaxation of best subset regression \citep{Tibshirani96}. The convexity
of (\ref{eqn-npmr}) makes it easier to solve than (\ref{eqn-rrmlm}), and we
discuss algorithms for solving it in Sections \ref{sub-pgd} and \ref{sub-apgd}.
In practice, we recommend using standard cross-validation techniques for
selecting the regularization parameter $\lambda$, which controls the rank of
the solution.

Consider the singular value decomposition $\MU^*\MSigma^*\MV^{*T}$ of the
$p \times K$
estimated coefficient matrix $\MB^*$. Each column of the $K\times K$ orthogonal
matrix $\MV^*$ represents a latent variable as a linear combination of the $K$
outcome categories. Meanwhile, each row of $\MU^*\MSigma^*$ specifies for each
predictor variable a coefficient for each {\em latent} variable, rather than
for each outcome category. By estimating some of the singular values of $\MB^*$
(the entries of the diagonal $p\times K$ matrix $\MSigma^*$) to be zero, we
reduce the number of coefficients to be estimated for each predictor variable
from (a) one for each of $K$ outcome categories; to (b) one for each of some
smaller
number of latent variables. These latent variables learned by the model express
relationships between the outcomes because two categories for which a latent
variable has both large positive coefficients are both likely to occur for
large values of that latent variable. Similarly, if a latent variable has a
large positive coefficient for one category and a large negative coefficient
for another, those two categories oppose each other diametrically with
respect to that latent variable.

\subsection{Proximal gradient descent}
\label{sub-pgd}

NPMR relies on solving
(\ref{eqn-npmr}). The objective is convex but non-differentiable where any
singular values of $\MB$ are zero, so we cannot use gradient descent.
Generally, when minimizing a function $f:\mR^d\rightarrow\mR$ of a vector
$x\in\mR^d$, the gradient descent update of step size $s$ takes the form

\begin{equation*}
x^{(t+1)} = x^{(t)} - s\nabla f(x^{(t)}),
\end{equation*}
or equivalently,

\begin{equation*}
x^{(t+1)} = \underset{x \in \mR^d}{\arg\min}\l\{f(x^{(t+1)}) + \langle \nabla
  f(x^{(t)}), x - x^{(t)}\rangle + \frac1{2s^{(t)}}||x - x^{(t)}||_2^2\r\}.
\end{equation*}

Still, if $f$ is non-differentiable, as it is in (\ref{eqn-npmr}), then
$\nabla f$ is undefined. However, if $f$ is the sum of two convex functions
$g$ and $h$, with $g$ differentiable, we can instead apply the generalized
gradient update step \citep{SLS}:

\begin{equation}
\label{eqn-generalized-gradient}
x^{(t+1)} = \underset{x \in \mR^d}{\arg\min}\l\{g(x^{(t+1)}) + \langle \nabla
  g(x^{(t)}), x-x^{(t)}\rangle + \frac1{2s^{(t)}}||x-x^{(t)}||_2^2 + h(x)\r\}.
\end{equation}

Repeatedly applying this update is known as proximal gradient descent (PGD).
In (\ref{eqn-npmr}), we have $x = \alpha, \MB$, $g = -\ell$ and
$h = ||\cdot||_*$. So the PGD update step is:

\begin{align*}
\alpha^{(t+1)},~\MB^{(t+1)} = \arg\min_{\alpha,~\MB}\l\{-\ell(\alpha^{(t)},\r.&
    \MB^{(t)}; \MX, \MY)\\
    + \langle \MX^T(\MY&- \MP^{(t)}), \MB - \MB^{(t)} \rangle
    + \langle 1_n^T(\MY - \MP^{(t)}), \alpha - \alpha^{(t)}\rangle\\
    +&\frac1{2s}||\MB - \MB^{(t)}||_F^2
    + \frac1{2s}\l. ||\alpha - \alpha^{(t)}||_2^2 + \lambda||\MB||_* \r\},
\end{align*}
where $\MY \in \{0, 1\}^{n\times K}$ is the matrix containing the response
variable and $\MP \in (0, 1)^{n\times K}$ is the matrix containing the fitted
values. That is, for $i = 1, ..., n$, and $k = 1, ..., K$,
\begin{equation}
\label{eqn-yp}
\{\MY\}_{ik} = \mI_{\{Y_i = k\}}, \hspace{5mm}\mbox{and}\hspace{5mm}
    \{\MP\}_{ik} = \frac{e^{\alpha_k + \vx_i^T\beta_k}}{\sum_{\ell=1}^K
    e^{\alpha_\ell + \vx_i^T\beta_\ell}}.
\end{equation}

The problem is separable in $\alpha$ and $\MB$:

\begin{align}
\begin{split}
\label{eqn-PGDint}
\alpha^{(t+1)} & = \arg\min_\alpha\l\{ \langle 1_n^T(\MY - \MP^{(t)}),
    \alpha-\alpha^{(t)}\rangle + \frac1{2s}||\alpha-\alpha^{(t)}||_2^2\r\}\\
    & = \alpha^{(t)} + s1_n^T(\MY - \MP^{(t)}), \mbox{ and}
\end{split}
\end{align}

\begin{align}
\begin{split}
\label{eqn-PGDmat}
\MB^{(t+1)}&=\arg\min_\MB\l\{\langle\MX^T(\MY-\MP^{(t)}),\MB-\MB^{(t)} \rangle
    + \frac1{2s}||\MB - \MB^{(t)}||_F^2 + \lambda||\MB||_*\r\}\\
    & = \mathcal S^*_{s\lambda}(\MB^{(t)} + s\MX^T(\MY - \MP^{(t)})),
\end{split}
\end{align}
where $\mathcal S^*_{s\lambda}:\mR^{p\times K} \rightarrow \mR^{p\times K}$ is
the soft-thresholding operator on the singular values of a matrix. Explicitly,
if a matrix $\MM\in\mR^{p\times K}$ has singular value decomposition
$\MU\MSigma\MV^T$, then
$\mathcal S^*_{s\lambda}(\MM) = \MU\mathcal S_{s\lambda}(\MSigma)\MV^T$, where
\begin{equation*}
\{\mathcal S_{s\lambda}(\MSigma)\}_{jk} = \mbox{sign}(\MSigma_{jk})
\max\{|\MSigma_{jk}| - s\lambda, 0\}.
\end{equation*}
$\mathcal S_{s\lambda}^*$ is called the {\it proximal operator} of the nuclear
norm,
and in general solving (\ref{eqn-generalized-gradient}) involves the proximal
operator of $h$, hence the name proximal gradient descent.

So to solve (\ref{eqn-npmr}), initialize $\alpha$ and $\MB$, and
iteratively apply the updates (\ref{eqn-PGDint}) and (\ref{eqn-PGDmat}). Due to
\cite{Nesterov07}, this procedure converges with step size $s \in (0, 1/L)$ if
the log-likelihood $\ell$ is continuously differentiable and has Lipschitz
gradient with Lipschitz constant $L$. The appendix includes a proof
that the gradient
of $\ell$ is Lipschitz with constant $L = \sqrt{K}||\MX||_F^2$, but in practice
we recommend starting with step size $s = 0.1$ and halving the step size if
any proximal gradient descent step would result in an increase of the objective
function (\ref{eqn-npmr}).

\subsection{Accelerated PGD}
\label{sub-apgd}

In practice, we find that it helps to speed things up considerably to use an
accelerated PGD method, also due to \cite{Nesterov07}. Specifically, we
iteratively apply the following updates:

\begin{enumerate}
\item $\alpha^{(t+1)} = \alpha^{(t)} +
  s1_n^T\l(\MY - \MP^{(t)}\r)$
\item $\MA^{(t+1)} = \MB^{(t)} + \frac t{t+3}(\MB^{(t)} - \MB^{(t-1)})$
\item $\MP^{(t+1)} = \MP(\alpha^{(t+1)}, A^{(t+1)})$
\item $\MB^{(t+1)} = \mathcal S^*_{s\lambda}\l(\MA^{(t+1)} +
  s\MX^T\l(\MY - \MP^{(t+1)}\r)\r)$
\end{enumerate}

The function $\MP(\cdot)$ in Step 3 returns the matrix of fitted probabilities
based on the regression coefficients as described in (\ref{eqn-yp}). Step 2 is
the key to the acceleration because it uses the ``momentum'' in $\MB$ to push
it further in the same direction it is heading. We strongly recommend using
this accelerated version of PGD, and
our implementation of NPMR is available on the Comprehensive R Archive Network
as the R package {\tt npmr}.

\subsection{Related work}
\label{sub-review}

The idea of using a nuclear norm penalty as a convex relaxation to reduced-rank
regression has previously been proposed
in the Gaussian regression setting \citep{Chen-etal13}, but we are not aware of
any attempt to do so in the multinomial setting.

The nearest competitor to NPMR that can feasibly be applied to the baseball
matchup dataset is multinomial ridge regression, which penalizes the squared
Frobenius norm (the sum of the squares of the entries) of the
coefficient matrix, instead of the nuclear norm. In detail, ridge
regression estimates the regression coefficients by solving the optimization
problem

\begin{equation}
\label{eqn-ridge}
(\alpha^*, \MB^*) = \underset{\alpha\in\mR^K,~B\in\mR^{p\times K}}{\arg\min}
    -\ell(\alpha^{(t)}, \MB^{(t)}; \MX, Y) + \lambda||\MB||_F^2.
\end{equation}

This model is very similar to the state of the art in public sabermetric
literature for evaluating pitchers on the basis of outcomes while
simultaneously controlling for sample size, opponent strength and ballpark
effects \citep{DRA}.
Software is available to solve this problem very quickly in the R package
{\tt glmnet} \citep{glmnet}. This is the standard approach used for regularized
multinomial regression problems, so we use it as the benchmark against which
to compare the performance of NPMR in Sections \ref{sec-sim} and
\ref{sec-results}.

\section{Simulation study}
\label{sec-sim}

In this section we present the results of two different simulations, one using
a full-rank coefficient matrix and the other using a low-rank coefficient
matrix. In both settings we vary the training sample size $n$ from 600 to 2000,
and we fix the number of predictor variables to be 12 and the number of
levels of the response variable to be 8. Given design matrix
$\MX \in \mR^{n\times12}$ and coefficient matrix $\MB \in \mR^{12\times8}$, we
simulate the response according to the multinomial regression model. Explicity,
for $i = 1, ..., n$, and $k = 1, ..., 8$,

$$\mP(Y_i = k) = \frac{e^{\MX\beta_k}}{\sum_{\ell = 1}^8e^{\MX\beta_\ell}}.$$

For both simulations the entries of $\MX$ are i.i.d. standard normal:

$$\vx_i\iid\mbox{Normal}(\vec0_{12}, I_{12})$$
for $i = 1, ..., n$. However the simulations differ in the generation of the
coefficient matrix $\MB$. In the {\em full rank} setting, the entries of $\MB$
follow an i.i.d. standard normal distribution: For $k = 1, ..., 8$,

\begin{equation}
\label{eqn-simulation}
\beta_k\iid\mbox{Normal}(\vec0_{12}, I_{12}).
\end{equation}

\begin{figure}[h]
	\centering
	\begin{subfigure}{.45\linewidth}
		\centering
		\includegraphics[width=\linewidth]{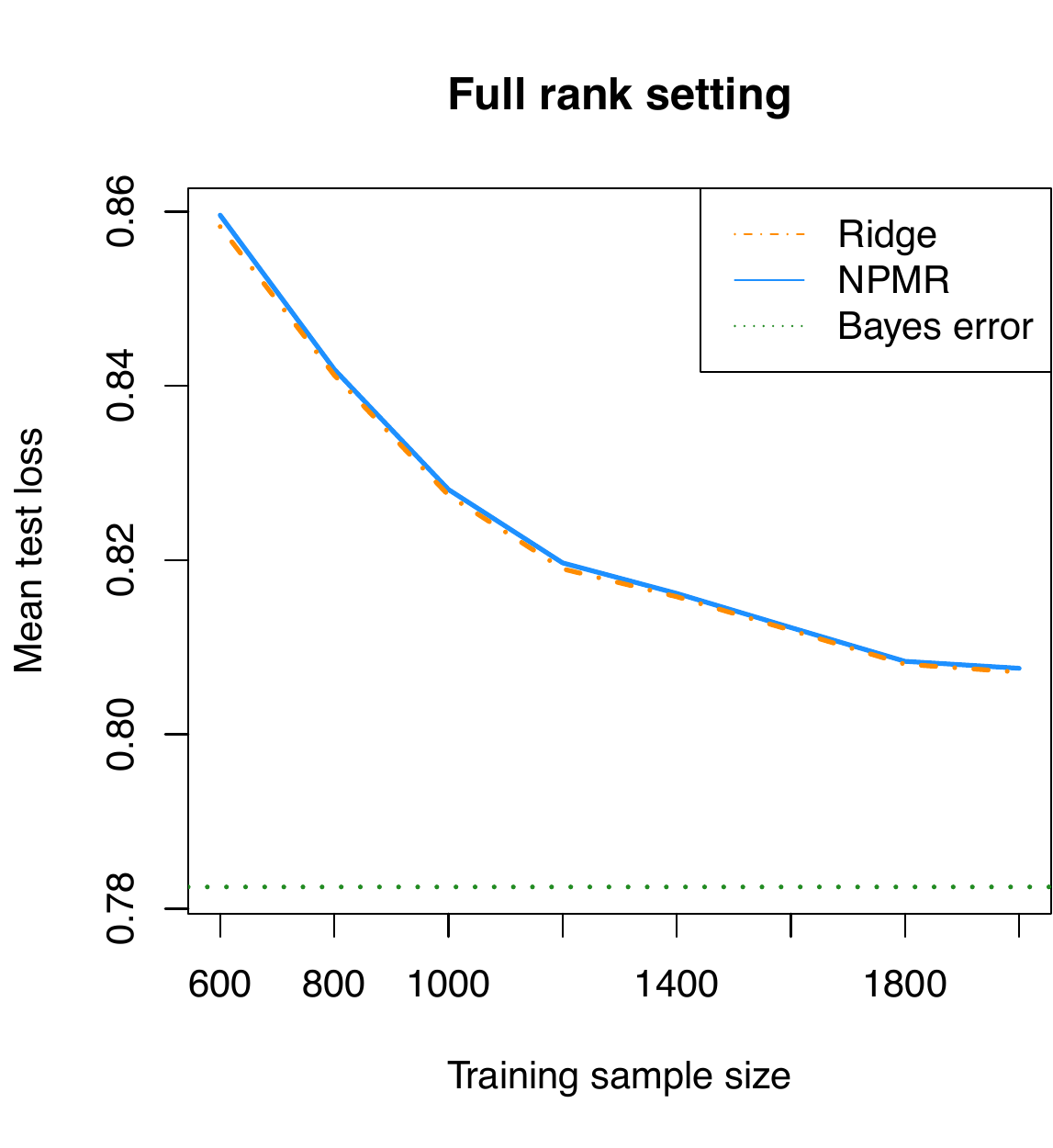}
		\caption{\it Full-rank coefficient matrix}
	\end{subfigure}
	\hspace{1cm}
	\begin{subfigure}{.45\linewidth}
		\centering
		\includegraphics[width=\linewidth]{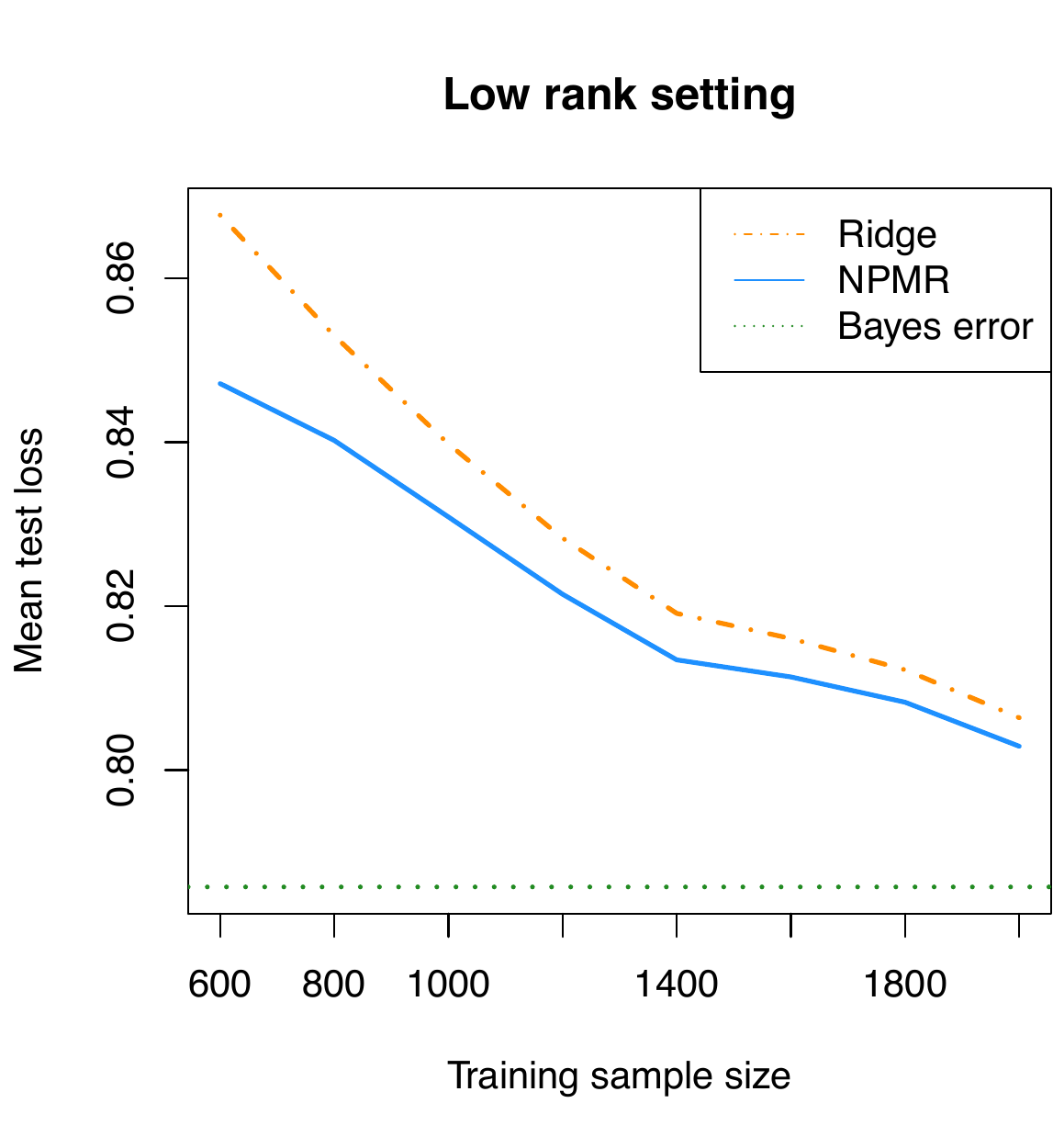}
		\caption{\it Low-rank coefficient matrix}
	\end{subfigure}
	\caption{\it Simulation results. We plot out-of-sample test error (using
        log-likelihood loss) against training sample size. The Bayes error is a
        lower bound on achievable test error.
        In (a), the full rank setting, ridge regression
        out-performs NPMR in terms of test error by a slim margin. In (b), the
        low rank setting, NPMR wins, especially for smaller sample sizes.}
	\label{fig-simulation}
\end{figure}

In the {\em low rank} setting we first simulate two intermediary matrices
$\MA\in\mR^{12\times2}$ and $\MC\in\mR^{8\times2}$ with i.i.d. standard normal
entries, and we then define $\MB\equiv\MA\MC^T$ so that the rank of $\MB$ is 2.
In each simulation we fit ridge regression and NPMR to the training sample of
size $n$ and estimate the out-of-sample error by simulating 10,000 test
observations, comparing the model's predictions on those test observations with
the simulated response. The results of 3,500 simulations in each setting, for
each training sample size $n$, are presented in Figure~\ref{fig-simulation}.

In the full rank setting we expect ridge regression to out-perform NPMR because
ridge regression shrinks all coefficient estimates toward zero, which is the
mean of the generating distribution for the coefficients in the simulation. If
this were a Gaussian regression problem instead of a multinomial regression
problem, then the ridge regression coefficient estimates would
correspond \citep{ESL} to the posterior mean estimate under a Bayesian prior of
(\ref{eqn-simulation}). In fact ridge regression does beat NPMR in this
simulation (for all training sample sizes $n$), but NPMR's performance is
surprisingly close to that of ridge regression.

The low rank setting is one in which NPMR should lead to a lower test error
than does ridge regression. NPMR bets on sparsity in the singular values
of the coefficient matrix, and in this setting the bet pays off. The simulation
results verify that this intuition is correct. NPMR beats ridge regression
for all training sample sizes $n$ but especially for smaller sample sizes. By
betting (correctly in this case) on the coefficient matrix having less than
full rank, NPMR learns more accurate estimates of the coefficient matrix.
As the training sample size increases, learning the
coefficient matrix becomes easier, and the performance gap between the two
methods shrinks but remains evident.

In summary, this simulation demonstrates that each of NPMR and ridge regression
is superior in a simulation tailored to its strengths,
confirming our intuition. Furthermore, in a simulation constructed in favor of
ridge regression, NPMR performs nearly as well. Meanwhile NPMR leads to more
significant gains over ridge regression in the low rank setting.

\section{Results}
\label{sec-results}

\subsection{Implementation details}

The 2015 MLB play-by-play dataset from Retrosheet includes
an entry for every plate appearance during the six-month regular season. For
the purposes of fitting NPMR to predict the outcomes of PAs, the following
relevant variables are recorded for the $i^{th}$ PA: the identity ($B_i$) of
the batter; the identity ($P_i$) of the pitcher; the identity ($S_i$) of the
stadium where the PA took place; an indicator ($H_i$) of whether the batter's
team is the home team; and finally an indicator ($O_i$) of whether the batter's
handedness (left or right) is opposite that of the pitcher.

For each outcome $k \in \K \equiv \{\mbox{K, G, F, BB, HBP, 1B, 2B, 3B, HR}\}$,
the multinomial model fit by both NPMR and ridge regression is specified by
\begin{align*}
\mP(Y_i = k) &= \frac{e^{\eta_{ik}}}{\sum_{\ell \in \K}e^{\eta_{i\ell}}},
    \mbox{ where}\\
\eta_{ik} &= \alpha_k + \beta_{B_ik} + \gamma_{P_ik} + \delta_{S_ik} +
    \zeta_kH_i + \theta_kO_i.
\end{align*}
The parameters introduced have the following interpretation: $\alpha_k$ is an
intercept corresponding to the league-wide frequency of outcome $k$;
$\beta_{B_ik}$ corresponds to the tendency of batter $B_i$ to produce outcome
$k$; $\gamma_{P_ik}$ corresponds to the tendency of pitcher $P_i$ to produce
outcome $k$; $\delta_{S_ik}$ corresponds to the tendency of stadium $S_i$ to
produce outcome $k$; $\zeta_k$ corresponds to the increase in likelihood of
outcome $k$ due to home field advantage; and $\theta_k$ corresponds to the
increase in likelihood of outcome $k$ due to the batter having the opposite
handedness of the pitcher's.

NPMR and ridge regression fit the same multinomial
regression model and differ only in the regularizations used in their objective
functions, yielding different results. See Section \ref{sec-npmr} for
details. However there is a minor tweak to NPMR for application to these data.
Instead of adding to the objective a penalty on the nuclear norm of the whole
coefficient matrix, we add penalties on the nuclear norms of the three
coefficient sub-matrices corresponding to batters, pitchers and stadiums. The
coefficients for home-field advantage and opposite handedness remain
unpenalized. The result is that NPMR learns different latent variables for
batters than it does for pitchers, instead of learning one set of latent
variables for both groups.

We process the PA data before applying NPMR and ridge regression. First, we
define a minimum PA threshold separately for batters and pitchers. For batters
the threshold is the $390^{th}$-largest number of PAs among all batters. This
corresponds roughly to the number of rostered batters at any given time during
the MLB regular season. Batters who fall below the PA threshold are labelled
``replacement level'' and within each defensive position are grouped together
into a single identity. For example, ``replacement-level catcher'' is a batter
in the dataset just like Mike Trout is, and the former label includes all PAs
by a catcher who does not meet the PA threshold. Similarly we define the
PA threshold for pitchers to be the $360^{th}$-largest number of PAs among all
pitchers, and we group all pitchers who fall below that threshold under the
``replacement-level pitcher'' label.
Additionally, we discard all PAs in which a pitcher is batting, and we discard
PAs which result in a catcher's interference or an intentional walk. The result
is a set of 176,559 PAs featuring 401 unique batters and 362 unique pitchers in
30 unique stadiums.

\subsection{Validation}

\begin{figure}[h!]
  \centering
  \includegraphics[width=.8\linewidth]{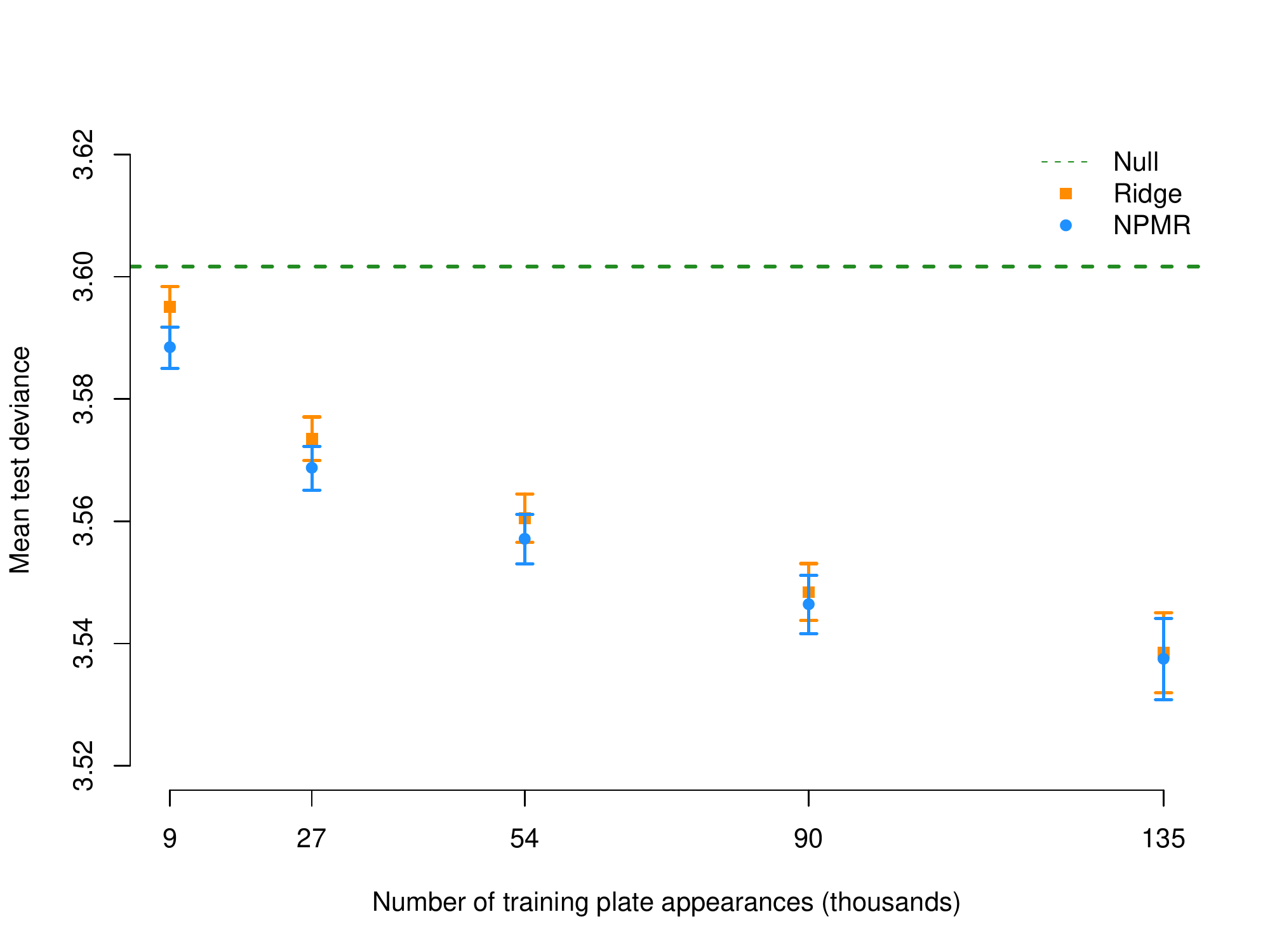}
  \caption{\it Out-of-sample test performance of NPMR, ridge and null
    estimators on
    baseball plate appearance result prediction. Each estimator was trained on
    a fraction of the 2015 regular season data (varying from 5 to 75 percent)
    and tested on the remaining data. The error bars correspond to one standard
    error.}
  \label{fig-baseball-test}
\end{figure}

We fit NPMR and ridge regression to the baseball data, using a training
sample that varied from 5\% (roughly 9,000 PAs) to 75\% (roughly 135,000 PAs)
of the data. In each case we used the remaining data to test the models, with
multinomial deviance (twice the negative log-likelihood) as the loss function.
For comparison we also include the null model, which predicts for each plate
appearance the league average probabilities of each outcome.
Figure~\ref{fig-baseball-test} gives the results.

For each training sample size, NPMR outperforms ridge regression though the
difference is not statistically significant. At the smallest sample size NPMR,
unlike ridge regression, achieves a significanly lower error than the null
estimator. There is value in improved estimation of players' skills in small
sample sizes because this can inform early-season decision-making. For all
other sample sizes, both NPMR and ridge regression achieves
errors which are statistically significantly less than the null error. The
primary benefit of NPMR relative to ridge regression is the interpretation, as
described in the next section.

\subsection{Interpretation}
\label{sub-interpret}

\begin{figure}[h]
  \centering
  \begin{subfigure}{\linewidth}
  	\centering
  	\includegraphics[width=.8\linewidth]{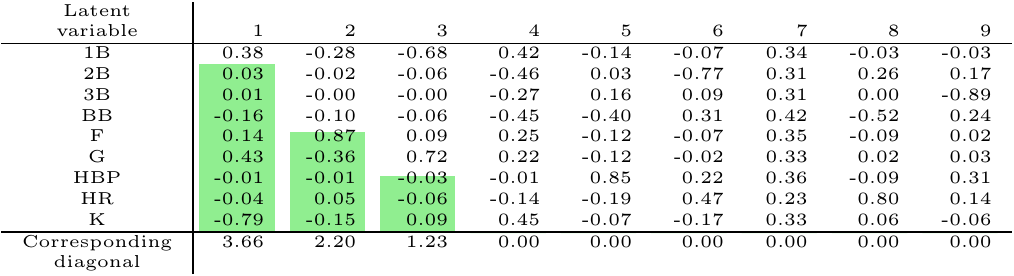}
  	\caption{\it Latent variables for batter regression coefficient matrix}
  \end{subfigure}
  \hspace{1cm}
  \begin{subfigure}{\linewidth}
  	\centering
  	\includegraphics[width=.8\linewidth]{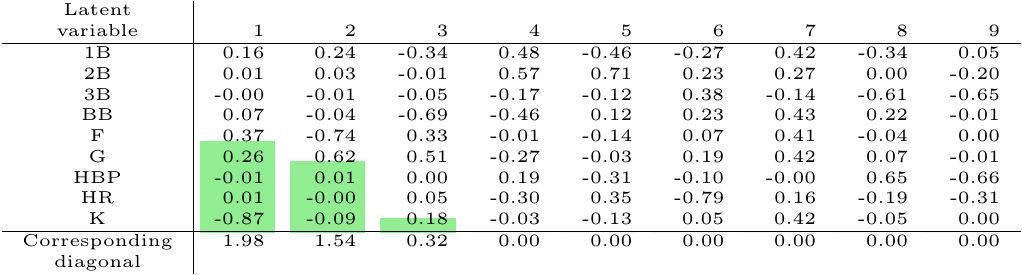}
  	\caption{\it Latent variables for pitcher regression coefficient matrix}
  \end{subfigure}
  \caption{\it Visualization of fitted regression coefficient matrices from
    NPMR on 5\% of the baseball data. The matrix displayed is $\MV$ in the
    $\MU\MSigma\MV^T$
    decomposition of $\MB$ from (\ref{eqn-npmr}), with columns
    corresponding to latent variables
    and rows corresponding to outcomes. The bottom row gives the entry in
    the diagonal matrix $\MSigma$ corresponding to the latent variable.}
  \label{fig-baseball}
\end{figure}

We focus on the results of fitting NPMR on 5 percent of the training data
because there the difference between NPMR and ridge regression is greatest
(Figure~\ref{fig-baseball-test}). As the sample size increases, the need for a
low-rank regression coefficient matrix is reduced, and the NPMR solution
becomes more similar to the ridge solution. Figure~\ref{fig-baseball}
visualizes the singular value decomposition of the fitted regression
coefficient submatrices corresponding to batters and pitchers.

We observe that for both batters and pitchers, NPMR identifies three latent
variables which differentiate players from one another. By construction,
these latent variables are measuring separate aspects of players' skills;
across players, expression in each latent skill is uncorrelated with
expression in each other latent skill. In that sense, we have identified three
separate skills which characterize
hitters and three separate skills which characterize pitchers. In baseball
scouting parlance, these skills are called ``tools'', but unlike the five
traditional baseball tools (hitting for power, hitting for contact, running,
fielding and throwing), the tools we identify are uncorrelated with one
another.

The interpretation of Figure~\ref{fig-baseball} is very attractive in the
context of domain knowledge. In reading the columns of $\MV$, note that they
are unique only up to a change in sign, so we can taken positive expression of
each skill to mean positive or negative values of the corresponding latent
variable. We suggest the following interpretation of the
first three latent skills for batters:

\begin{itemize}
  \item {\em Skill 1: Patience}. The loadings of the first
    latent variable discriminate perfectly between the TTO outcomes and the
    BIP outcomes described in Section \ref{sec-intro}. We label this skill as
    ``patience'' because when a batter swings at fewer pitches, he is less
    likely to hit the ball in play.
  \item {\em Skill 2: Trajectory}. The second latent variable distinguishes
    primarily between F and G, corresponding to the vertical launch angle of
    the ball off the bat.
  \item {\em Skill 3: Speed}. The third latent
    variable distinguishes primarily between 1B and G. Examining the players
    with strong positive expression of this skill, we find fast players who are
    more difficult to throw out at first base on a ground ball.
\end{itemize}

From this interpretation we learn that the primary skill which distinguishes
betwen batters is how often they hit the ball into the field of play. One
outcome over which batters have relatively large control is how often they
swing at pitches. Among balls that are put into play, batters have less but
still subtantial control over whether those are ground balls or fly balls. It
is the vertical angle of the batter's swing plane, along with whether he tends
to contact the top half or the bottom half of the ball, that determines his
trajectory tendency. Finally, given the trajectory of the ball off the bat, the
batter has relatively little control over the outcome of the PA. But to the
extent that he can influence this outcome, fast runners tend to hit more
singles and fewer groundouts.

Based on Figure~\ref{fig-baseball}, we interpret the pitchers' skills as
follows:

\begin{itemize}
  \item {\em Skill 1: Power}. The first latent variable distinguishes
    primarily between K and F (and G), thus identifying how the pitcher
    gets outs. Pitchers who tend to get their outs via the strikeout are
    known in baseball as ``power pitchers''.
  \item {\em Skill 2: Trajectory}. As with batters, the second latent variable
    distinguishes primarily between F and G, corresponding to the trajectory of
    the ball off the bat.
  \item {\em Skill 3: Command}. The third latent variable
    distinguishes primarily between positive outcomes for the pitcher
    (F, G and K) and negative outcomes for the pitcher (BB and 1B), reflecting
    how well is able to control the location of his pitches.
\end{itemize}

The interpretation of the first two skills for pitchers is very similar to the
interpretation of the first two skills for batters. Primarily, pitchers can
influence how often balls are hit into play against them, but they exhibit less
control over this than batters do. Secondarily, as with hitters, pitchers
exhibit some control over the vertical launch angle of the ball off the bat.
This is based on the location and movement of their pitches. The third skill,
distinguishing between positive and negative outcomes, has a relatively very
small magnitude.

\begin{table}[h]
\caption{\it Top 5 and bottom 5 batters in the three latent skills
    identified by NPMR.}
\centering
{\tiny\begin{tabular}{c|c|c|c|c|}
\hline
{\bf Skill}       &
{\bf Patience}    & {\bf Trajectory}  & {\bf Speed}\\
\hline
\hline
&More K, BB       & More F              & More 1B\\
\hline
& Peter Bourjos   & Ian Kinsler         & Yoenis Cespedes\\
Top
& Eddie Rosario   & Freddie Freeman     & Lorenzo Cain\\
5
& Carlos Santana  & Omar Infante        & Jos\'{e} Iglesias\\
& George Springer & Kolten Wong         & Kevin Kiermaier\\
& Mike Napoli     & Jos\'{e} Altuve     & Delino DeShields Jr\\
&                 &                     & \\
& Josh Reddick    & Dee Gordon          & Evan Longoria\\
Bottom
& JT Realmuto     & Alex Rodriguez      & Ryan Howard\\
5
& AJ Pollock      & Cameron Maybin      & Odubel Herrera\\
& Kevin Pillar    & Shin-Soo Choo       & Seth Smith\\
& Eric Aybar      & Francisco Cervelli  & Jake Lamb\\
\hline
&More F, G, 1B    & More G, 1B          & More G\\
\hline
\end{tabular}}
\label{tab-batters}
\end{table}

Table \ref{tab-batters} lists the top five and bottom five players in each of
the three latent batting skills learned by NPMR. These results largely match
intuition for the players listed, and to the extent that they do not, it is
worth a reminder that they are based on roughly nine days' worth of data from a
six-month season. The median number of PAs per batter in the
training set is 21.

\begin{table}[h]
\caption{\it Top 5 and bottom 5 pitchers in the three latent skills
    identified by NPMR.}
\centering
{\tiny\begin{tabular}{c|c|c|c|}
\hline
{\bf Tool}          &
{\bf Power}         & {\bf Trajectory}  & {\bf Command}\\
\hline
\hline
&More K             & More F            & More F, G, K\\
\hline
&Jos\'{e} Quintana  & Jesse Chavez      & Max Scherzer\\
Top
&Corey Kluber       & Justin Verlander  & Masahiro Tanaka\\
5
&Madison Bumgarner  & Jake Peavy        & Jacob deGrom\\
&Max Scherzer       & Johnny Cueto      & Rubby de la Rosa\\
&Clayton Kershaw    & Chris Young       & Matt Harvey\\
&                   &                   & \\
&John Danks         & Dallas Keuchel    & Mike Pelfrey\\
Bottom
&Dan Haren          & Garrett Richards  & Chris Tillman\\
5
&Cole Hamels        & Sam Dyson         & Eddie Butler\\
&Alfredo Sim\'{o}n  & Brett Anderson    & Gio Gonzalez\\
&RA Dickey          & Michael Pineda    & Jeff Samardzija\\
\hline
&More F, G          & More G            & More BB, 1B\\
\hline
\end{tabular}}
\label{tab-pitchers}
\end{table}

The results in Table \ref{tab-pitchers}, listing the top
and bottom players in each of the three latent pitching skills, are more
interesting. The top five power pitchers are all among to top starting pitchers
in the game. All the way on the other side of the spectrum is knuckleball
pitcher RA Dickey. The knuckleball is a unique pitch in baseball thrown
relatively softly with as little spin as possible to create unpredictable
movement. Its goal is not to overpower the opposing batter but to induce weak
contact. Another interesting pitcher low on power is Cole Hamels. Two of the
leading sabermetric websites, Baseball Prospectus and FanGraphs, disagree
greatly on Hamels' value. The discrepancy stems from Baseball Prospectus giving
full weight to BIP outcomes while FanGraphs ignores them. Because Hamels tends
to get outs via fly balls and ground balls rather than strikeouts, FanGraphs
estimates a much lower value for Hamels than Baseball Prospectus does.

\subsection{Another application: Vowel data}

Consider the problem of vowel classification from \cite{Robinson89}. The data
set comprises 528 training samples and 462 test samples, each classified as one
of the 11 vowels listed in Table \ref{tab-vowels}, with 10 features extracted
from an audio file. The data are grouped by speaker, with 8 subjects in the
training set and 7 different subjects in the test set. Each audio clip is split
into 6 frames during a duration of steady audio, yielding 6 pseudo-replicates.

\begin{table}[h]
\caption{\it Symbols and words for vowels studied by \cite{Robinson89}.}
\label{tab-vowels}
\centering
\begin{tabular}{cc|cc|cc|cc}
Vowel   & Word  & Vowel & Word  & Vowel & Word  & Vowel & Word\\
\hline
i       & heed  & A     & had   & O     & hod   & u:    & who'd\\
I       & hid   & a:    & hard  & C:    & hoard & 3:    & heard\\
E       & head  & Y     & hud   & U     & hood\\
\end{tabular}
\end{table}

\begin{figure}[h]
\caption{\it Results of fitting NPMR and ridge regression on vowel data. Test
    error is plotted against training error, using negative log-likelihood loss.
    Training error serves as a surrogate for degrees of freedom in the model
    fit. The null prediction assigns equal probability to all categories. Error
    bars represent one standard error in estimation of the test loss.}
\includegraphics[width = 0.7\textwidth]{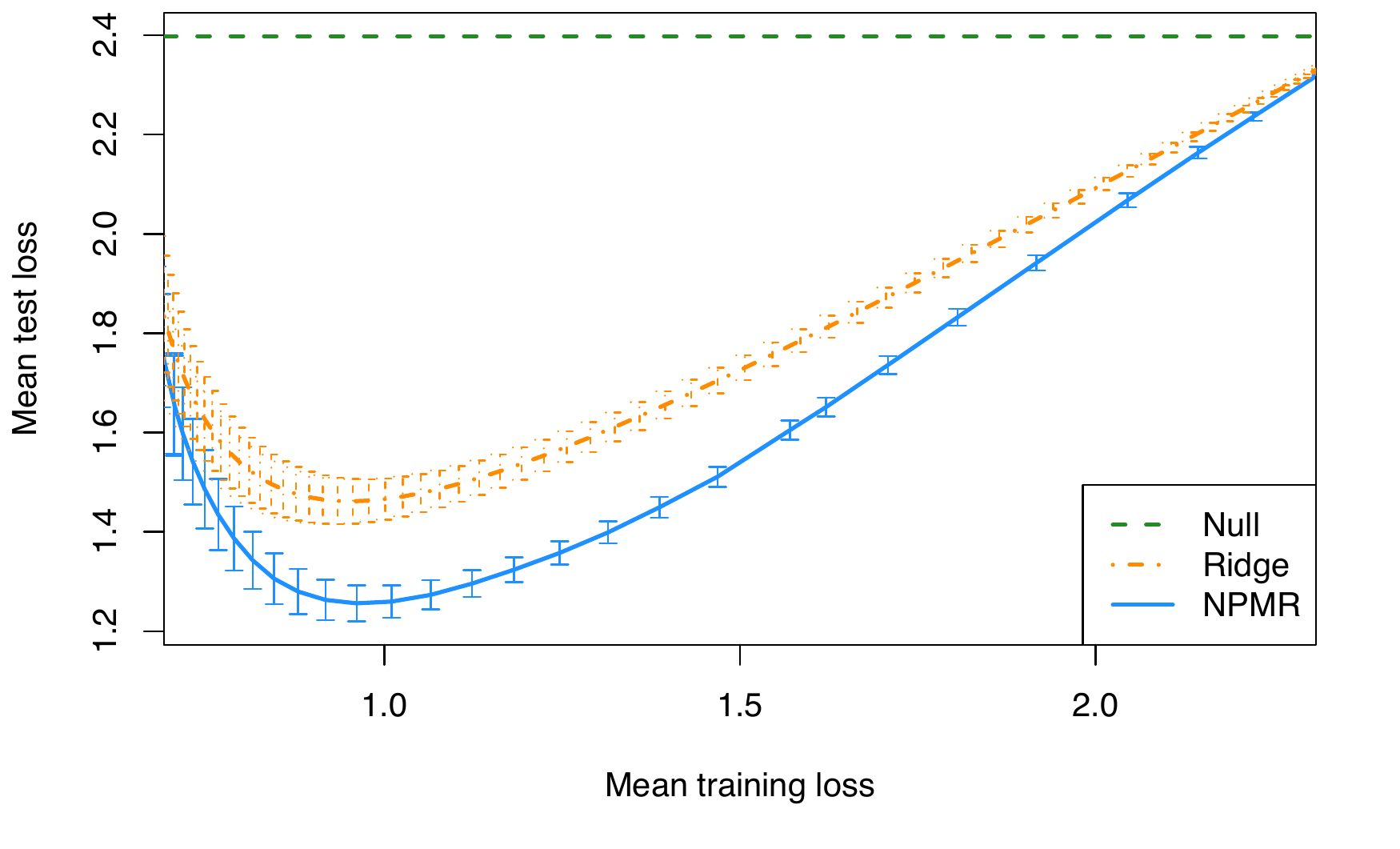}
\label{fig-vowel-results}
\end{figure}

We fit NPMR and ridge regression to the training data over a wide range
of regularization parameters, with the training and test loss (negative
log-likelihood)
reported in Figure~\ref{fig-vowel-results}. As the regularization parameter
increases for each method, the training loss increases. The test loss initially
decreases and then increases as the model is overfit. We observe that over
the whole solution path, for the same training error NPMR consistently yields a
lower test error than ridge regression.

\begin{figure}[h]
\caption{\it Visualization of fitted regression coefficient matrices from NPMR
        applied to the vowel data. 
        The matrix displayed is $\MV$ in the $\MU\MSigma\MV^T$ decomposition of
        the regression coefficient matrix $\MB$,
        with columns corresponding to latent variables and rows corresponding
        to outcomes. The bottom row gives the entry in the diagonal matrix
        $\MSigma$ corresponding to the latent variable.}
\includegraphics[width = \textwidth]{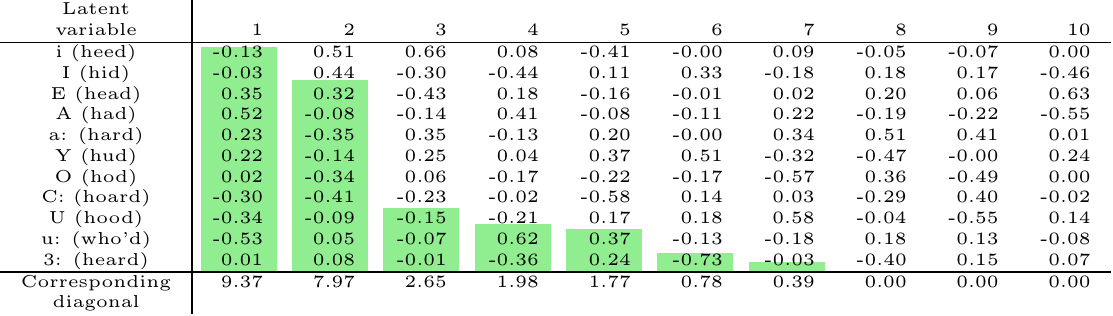}
\label{fig-vowel-B}
\end{figure}

Figure~\ref{fig-vowel-B} reveals a possible explanation why NPMR outperforms
ridge regression on the
vowel data. For example the results show that when the vowel i is a likely
label, the vowel I is also a likely label. The first two latent variables
explain a significant portion of the
variance in the regression coefficients for the vowels.
The first latent variable distinguishes between two groups of vowels,
with C:, U and u: having the most negative values and E, A, a: and Y having the
most positive values. NPMR has beaten ridge regression here by leveraging a
hidden structure among response categories.

\section{Discussion}
\label{sec-discussion}

The potential for reduced-rank multinomial regression to leverage the
underlying structure among response categories has been recognized in the past.
But the computational cost for the state-of-the-art algorithm for fitting such
a model is so great as to make it infeasible to apply to a dataset as large as
the baseball play-by-play data in the present work. Using a
convex relaxation of the problem, by penalizing the nuclear norm of the
coefficient matrix instead of its rank, leads to better results.

The interpretation of the results on the baseball data is promising in how it
coalesces with modern baseball understanding. Specifically,
the NPMR model has quantitative implications on leveraging the structure in PA
outcomes to better jointly estimate outcome probabilities. Additional
application to vowel recognition in speech shows improved out-of-sample
predictive performance,
relative to ridge regression. This matches the intuition that NPMR is
well-suited to multinomial regression in the presence of a generic structure
among the response categories. We recommend the use of NPMR for any multinomial
regression problem for which there is some nonordinal structure among the
outcome categories.

\section*{Acknowledgments}
The authors would like to thank Hristo Paskov, Reza Takapoui and Lucas Janson
for helpful discussions, as well as Balasubramanian Narasimhan for
computational assistance.
% and are grateful to an
% Editor and an Associate Editor for helpful comments that led to improvements
% to this work.

\bibliographystyle{apalike}
\bibliography{powers}

\newpage

\appendix

\section{Appendix}

\subsection*{Identifiability of multinomial logistic regression model}

We observe in Section \ref{sec-rrmlm} that the model (\ref{eqn-multinomial}) is
not identifiable: For any $a \in \mR$ and $\vc \in \mR^p$,

\begin{equation*}
\frac{e^{\alpha_k - a + \vx_i^T(\beta_k - \vc)}}
  {\sum_{\ell = 1}^Ke^{\alpha_\ell - a + \vx_i^T(\beta_\ell - \vc)}} =
\frac{e^{-a - x_i^T\vc}e^{\alpha_k + \vx_i^T\beta_k}}
  {e^{-a - x_i^T\vc}\sum_{\ell = 1}^Ke^{\alpha_\ell + \vx_i^T\beta_\ell}} =
\frac{e^{\alpha_k + \vx_i^T\beta_k}}
  {\sum_{\ell = 1}^Ke^{\alpha_\ell + \vx_i^T\beta_\ell}}
\end{equation*}

Hence $(\alpha, \MB)$ and $(\alpha - a1_K, \MB - \vc1_K^T)$ have the same
likelihood. The ridge penalty in (\ref{eqn-ridge}) provides a natural
resolution. Any solution to this problem must satisfy

\begin{equation}
\label{eqn-ridge-id}
||\MB||_F^2 = \min_{\vc \in \mR^p}||\MB - \vc1_K^T||_F^2
\end{equation}
because otherwise $\MB$ can be replaced by $\MB - \vc1_K^T$ with a smaller
norm but the same likelihood and hence a lesser objective. Note that the
optimization problem on the right-hand side of
(\ref{eqn-ridge-id}) is separable in the entries of $\vc$ and has the unique
solution $\vc^* = \frac1K\MB1_K$, meaning that the rows of $\MB$ in the
solution must have mean zero. The unpenalized intercept $\alpha$ stil lacks
identifiability, but we may take it to have mean zero as well.

Similarly, the NPMR solution must satisfy
\begin{equation}
\label{eqn-npmr-id}
||\MB||_* = \min_{\vc \in \mR^p}||\MB - \vc1_K^T||_*.
\end{equation}
Whether this optimization problem always (for any $\MB \in \mR^{p\times K}$)
has a unique solution is an open question. We speculate that it does and that
the unique solution is $\vc^* = \frac1K\MB1_K$. As evidence, each fit of NPMR
in the present manuscript has a solution with zero-mean rows. As further
evidence, we have used the MATLAB software CVX \citep{cvx} to solve
(\ref{eqn-npmr-id}) for several randomly generated matrices $\MB$, and each
time the solution has been $\vc^* = \frac1K\MB1_K$.

Note that $\vc^* = \frac1K\MB1_K$ must always be {\it a} solution to
(\ref{eqn-npmr-id}). To see this, note that
\begin{equation*}
\MB - \vc^*1_K^T = \MB - \frac1K\MB1_K1_K^T = \MB\l(\MI - \frac1K1_K1_K^T\r) =
\MB(\MI - \MH),
\end{equation*}
where $\MH = 1_K(1_K^T1_K)^{-1}1_K^T$ is a projection matrix. Hence $\MI - \MH$
is also a projection matrix and has spectral norm (maximum singular value)
${||\MI - \MH||_\infty = 1}$. By H\"older's inequality for Schatten $p$-norms
\citep{MatrixAnalysis},

\begin{equation*}
||\MB(\MI - \MH)||_* \le ||\MB||_*||\MI - \MH||_\infty = ||\MB||_*,
\end{equation*}
so for any $\MB \in \mR^{p\times K}$, $$||\MB - \frac1K\MB1_K1_K^T||_* \le
||\MB||_*.$$
In order words, the nuclear norm can always be decreased, or at least not
increased, by centering the rows to have mean zero.

The problem with a lack of identifiability in the multimonial regression model
comes in the interpretation of the regression coefficients. When comparing
coefficients across variables for the same outcome class, it is concerning that
an arbitrary increase in either coefficient can corresond to the same fitted
probabilities (if that same increase applys to all other coefficients for the
same variable). This does not apply to any of the interpretation in Section
\ref{sub-interpret}, but in the absence of certainty that there is a unique
solution to (\ref{eqn-npmr-id}), we take the NPMR solution to be the one for
which the mean of $\alpha$ and the row means of $\MB$ are zero.

\subsection*{Proof of Lipschitz condition for multinomial log likelihood}

We prove that the multinomial log-likelihood $\ell(\alpha, \MB; \MX, Y)$ from
(\ref{eqn-npmr}) has Lipschitz gradient with constant
$L = \sqrt{K}||\MX||_F^2$. Assume (without loss of generality) that the
covariate matrix $\MX$ has a column of 1s encoding the intercept, so
$\alpha = 0$. The gradient of $\ell(\MB;\MX,Y)$ with respect to $\MB$ is given
by $\MX^T(\MY - \MP)$, where $\MY$ and $\MP$ are defined as in (\ref{eqn-yp}).
What we must show is that, for any $\MB, \MB' \in \mR^{p\times K}$:
\begin{equation}
\label{eqn-lipschitz}
||\MX^T(\MY - \MP) - \MX^T(\MY - \MP')||_F \le
    \sqrt{K}||\MX||_F^2||\MB - \MB'||_F.
\end{equation}
Recall that $\MP$ is a function of $\MB$, so $\MP'$ corresponds to $\MB'$.

Consider a single entry $\MP_{ik}$ of $\MP$. Note that the gradient of
$\MP_{ik}$ with respect to $\MB$ is given by $x_iw_{ik}^T$, where
$w_{ik} \in \mR^p$ and
$$(w_{ik})_j = \l\{
    \begin{array}{cc}
    -\MP_{ik}\MP_{ij}         & j\ne k\\
    \MP_{ik}(1-\MP_{ik})    & j=k
    \end{array}
    \r..$$
For any $\MP \in (0, 1)^{n\times K}$,
$$||w_{ik}||_2 \le ||w_{ik}||_1 =
    \MP_{ik}(1-\MP_{ik}) + \MP_{ik}\sum_{j\ne k}\MP_{jk} =
    2\MP_{ik}(1-\MP_{ik}) \le \frac12.$$
This implies that the norm of the gradient of $\MP_{ik}$ is bounded above by
the inequality $||x_iw_{ik}^T||_F \le ||x_i||_2||w_{ik}^T||_F \le ||x_i||_2$.
So for any $\MB, \MB' \in \mR^{p\times K}$:
\begin{equation}
\label{eqn-inequality}
|\MP_{ik} - \MP_{ik}'| \le ||x_i||_2||\MB - \MB'||_F.
\end{equation}

Now we are ready to prove (\ref{eqn-lipschitz}).
\begin{align*}
||\MX^T(\MY - \MP) - \MX^T(\MY - \MP')||_F & = ||\MX^T(\MP - \MP')||_F\\
    & \le ||\MX||_F||\MP - \MP'||_F\\
    & = ||\MX||_F\sqrt{\sum_{i=1}^n\sum_{k=1}^K(\MP_{ik} - \MP_{ik}')^2}\\
    & \le ||\MX||_F\sqrt{\sum_{i=1}^n\sum_{k=1}^K||x_i||_2^2||\MB-\MB'||_F^2}
        \hspace{1cm}\mbox{from (\ref{eqn-inequality})}\\
    & = ||\MX||_F\sqrt{K||\MB-\MB'||_F^2\sum_{i=1}^n||x_i||_2^2}\\
    & = ||\MX||_F\sqrt{K||\MB-\MB'||_F^2||\MX||_F^2}\\
    & = \sqrt{K}||\MX||_F^2||\MB-\MB'||_F\hspace{1cm}\blacksquare\\
\end{align*}

\end{document}